\documentclass[10pt,twocolumn,letterpaper]{article}

\usepackage[pagenumbers]{cvpr} 

\usepackage{float}
\usepackage{newfloat}
\usepackage{listings}
\usepackage{subfiles} 
\usepackage{subcaption}
\usepackage{ragged2e} 
\usepackage{booktabs, makecell, multirow, tabularx}
\usepackage{bbding}
\usepackage{xcolor}
\usepackage{mathtools}
\usepackage{mathrsfs}
\usepackage{geometry} 
\usepackage{diagbox}

\usepackage{wrapfig}
\usepackage{floatflt}
\usepackage{listings}

\definecolor{cvprblue}{rgb}{0.21,0.49,0.74}
\usepackage[pagebackref,breaklinks,colorlinks,allcolors=cvprblue]{hyperref}

\def\paperID{40415} 
\def\confName{CVPR}
\def\confYear{2026}

\definecolor{lightgray}{gray}{0.9}
\newcommand{\model}{M$^3$D-BFS }

\title{M$^3$D-BFS: a Multi-stage Dynamic Fusion Strategy for Sample-Adaptive Multi-Modal Brain Network Analysis}

\author{
    Rui Dong, Xiaotong Zhang, Jiaxing Li, Yueying Li, Jiayin Wei, Youyong Kong\thanks{Corresponding author}
    \\
    \small School of Computer Science and Engineering, Southeast University
    \\
    \tt \small dongrui$\_$0427.seu.edu.cn, zxt$\_$000902@163.com, Jiaxing$\_$Li@seu.edu.cn,
    \\
    \tt \small yueyli@163.com, 220236174@seu.edu.cn, kongyouyong@seu.edu.cn    
}

\begin{document}

\maketitle

\begin{abstract}
    Multi-modal fusion is of great significance in neuroscience which integrates information from different modalities and can achieve better performance than uni-modal methods in downstream tasks. Current multi-modal fusion methods in brain networks, which mainly focus on structural connectivity (SC) and functional connectivity (FC) modalities, are static in nature. They feed different samples into the same model with identical computation, ignoring inherent difference between input samples. This lack of sample adaptation hinders model's further performance. To this end, we innovatively propose a multi-stage dynamic fusion strategy (M$^3$D-BFS) for sample-adaptive multi-modal brain network analysis. Unlike other static fusion methods, we design different mixture-of-experts (MoEs) for uni- and multi-modal representations where modules can adaptively change as input sample changes during inference. To alleviate issue of MoE where training of experts may be collapsed, we divide our method into 3 stages. We first train uni-modal encoders respectively, then pretrain single experts of MoEs before finally finetuning the whole model. A multi-modal disentanglement loss is designed to enhance the final representations. To the best of our knowledge, this is the first work for dynamic fusion for multi-modal brain network analysis. Extensive experiments on different real-world datasets demonstrates the superiority of M$^3$D-BFS.
    Code can be available at https://github.com/KamonRiderDR/M3D-BFS.
    
\end{abstract}

\begin{figure}
    \centering
    \includegraphics[width=0.96\linewidth]{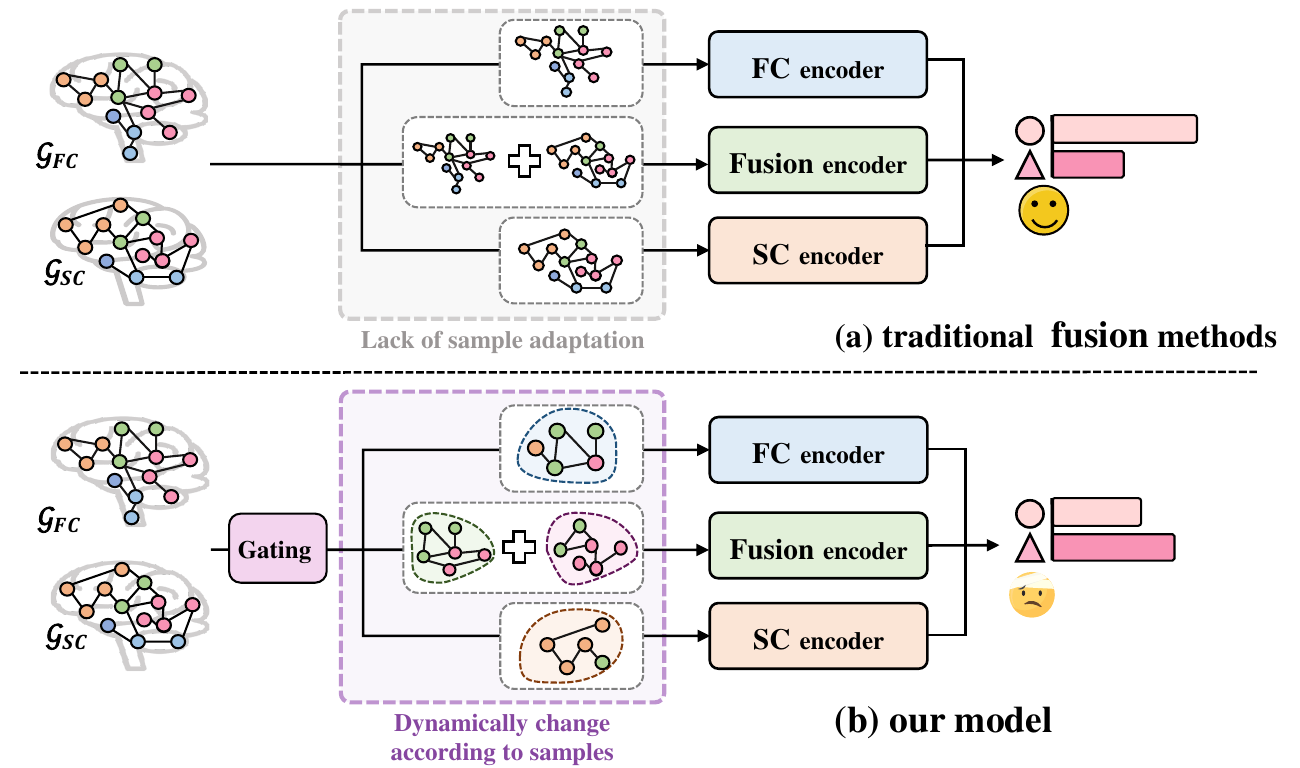}
    \caption{Illustration of M$^3$D-BFS. Compared with traditional methods, M$^3$D-BFS can adaptively adjust for different input samples.}
    \label{pdf_toy_illustration}
\end{figure}


\begin{figure*}[!tbp]
    \centering
    \includegraphics[width=0.92\textwidth]{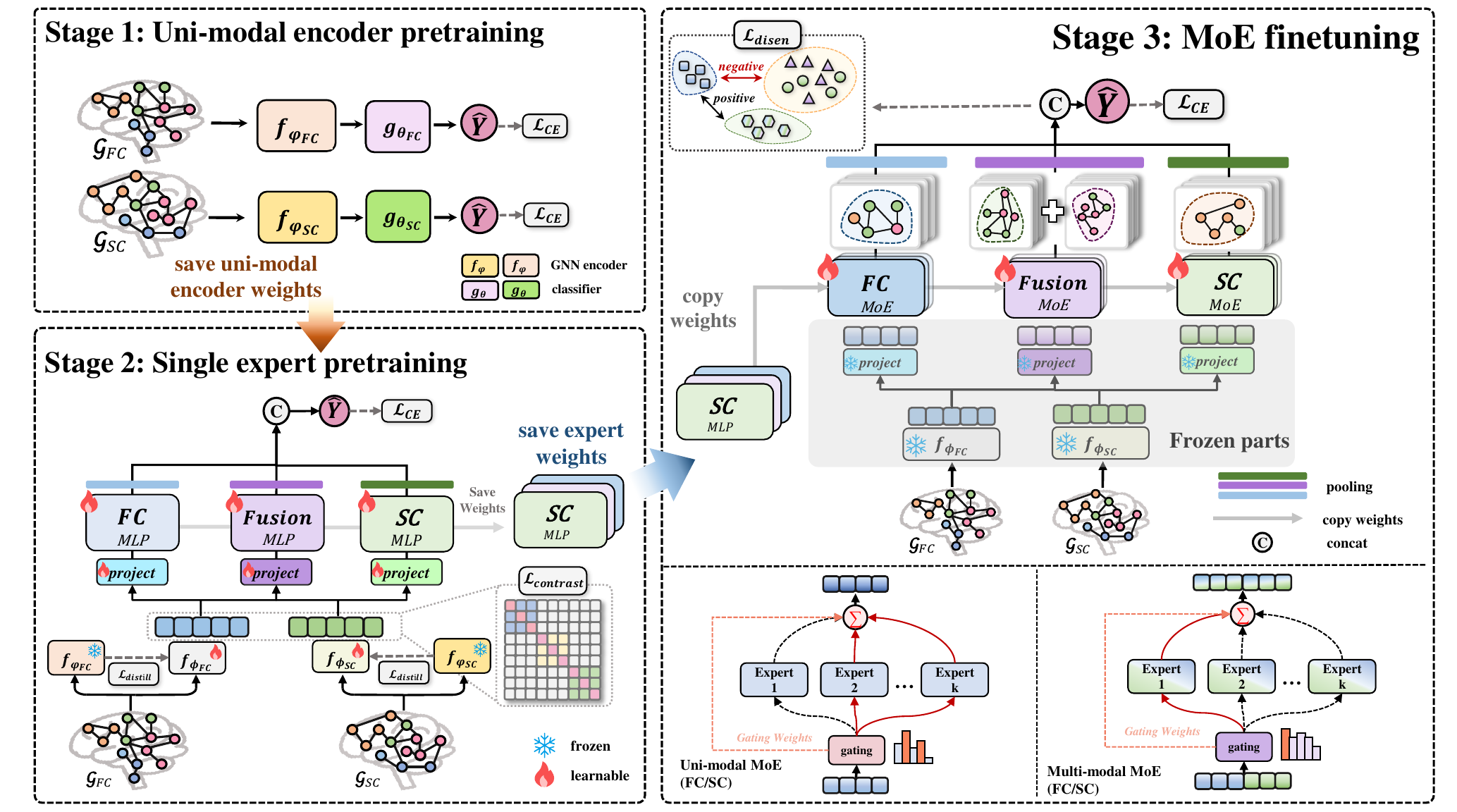}
    \caption{The overall framework of our proposed M$^3$D-BFS. It consists of 3 stages: \textbf{(1) stage 1}: training for uni-modal encoders. \textbf{(2) stage 2}: pretraining of M$^3$D-BFS by using single expert. \textbf{(3) stage 3}: Finetuning M$^3$D-BFS via finetuing MoE blocks. $\mathcal{L}_{disen}$ is designed to enhance final embeddings. Structures of MoEs (uni/multi-modal) are also plotted.}
    \label{M3D-BFS}
\end{figure*}

\section{Introduction}
Most current neuroscience studies are based on uni-modal learning, involving only one modality \cite{bessadok2022graph,zahid2022brainnet,winter2024systematic}. In brain network analysis, most works focus on functional magnetic resonance imaging (fMRI) \cite{yan2019reduced} and diffusion magnetic resonance imaging (dMRI) \cite{van2020white}. However, in existing real-world medical scenarios, data from SC and FC offer complementary information \cite{chu2018function}, while single-modality-based methods are insufficient to capture comprehensive useful messages. On the other hand, multi-modal fusion, which integrates information from different modalities, can utilize data more efficiently and achieve better performance in downstream tasks, like gender classification and disorder classification.

Multi-modal brain network fusion typically utilizes data from fMRI (functional) and dMRI (structural) modalities. Different from other image-based medical fusion methods, multi-modal brain network fusion mainly considers fusion between different graphs (FC and SC) to better capture topology features from brains. In general, current fusion stategies can be divided into 3 categories: \textbf{(a) Representation-level fusion} which directly concats multi-modal representations for fusion. \cite{sebenius2021multimodal,zhang2021deep,wen2024d}. \textbf{(b) Graph-level fusion} between multi-modal brain networks. ALNEGAT \cite{chen2022adversarial} designs an attention-based adjacency matrix for fusion. Cross-GNN \cite{yang2023mapping} proproses a graph-based learning framework to capture finer inter-modal dependencies. \textbf{(c) transformer-based} fusion methods. RH-BrainFS \cite{ye2024rh} considers regional heterogeneity between SC-FC and utilized bottleneck transformer for indirect fusion. NeuroPath \cite{wei2024neuropath} adds coupling mechanism between FC and SC to improve FC's performance.

Despite the advancements in multi-modal brain network fusion, current methods are static in nature. They feed all the inputs into the same models with identical computation, ignoring the inherent difference between different input samples. In Computer Vision and Natural Language Processing, this is called multi-modal dynamic fusion, which aims to adaptively fuse multi-modal data and make predictions based on the data-wise models \cite{Cao_2023_ICCV,han2022multimodal,xue2023dynamic}. Such issue also exists in multi-modal brain network analysis, where different samples may have significant differences in strength and connectivity of some brain regions \cite{basaia2024brain,duan2017degree,chou2023functional}. While current methods treat all the samples equally by using same models to encode and fuse, they fail to avoid the effects for inherent variance among different samples during encoding and fusion. In other word, \textit{model will have better performance if it can adaptively select unimodal-wise parts and multimodal-wise parts for representation and fusion as the input sample dynamically changes.} 

To fill the gap mentioned above in neuroscience, in this work we \textbf{innovatively propose a dynamic fusion approach for sample-adaptive multi-modal brain network analysis}. As is shown in Fig. \ref{pdf_toy_illustration}, for other traditional fusion methods, all the input will be directly sent into specific or fusion modules. Our method instead adds gating module before to adaptively capture modal-specific and multi-modal data for further representation. We realize our gating module based on mixture-of-experts (MoE). However, directly adopting MoE is a challenge as it may face expert collapse in training. Therefore, we propose a \underline{\textbf{M}}ulti-stage, \underline{\textbf{M}}ulti-\underline{\textbf{M}}odal \underline{\textbf{D}}ynamic-\underline{\textbf{B}}rain \underline{\textbf{F}}u\underline{\textbf{S}}ion Strategy (\textbf{M$^3$D-BFS}). \model consists of 3 stages: We first train uni-modal encoders for SC/FC modality respectively. Then we pretrain our model by training single experts for each MoE respectively. SC-FC contrastive loss and unimodal-guided distillation loss are designed for better performance of pretraining. Finally we finetune our model by only keeping MoE modules trainable. The weights for experts in previous stage are used to initialize experts in corresponding MoEs. To further enhance final representations (uni-modal/multi-modal) and their correlations with downstream tasks, we proposed a multi-mdoal disentangle loss to better capture task-related information. To the best of our knowledge, \model is the first work for dynamic fusion in sample-wise multi-modal brain network analysis. Extensive experiments on different real-world datasets demonstrate the effectiveness of our method. 

\section{Related Works}

\noindent\textbf{Dynamic Multi-modal Fusion.} Firstly introduced in Computer Vision, dynamic fusion aims to dynamically select parts from multi-view images for fusion according to different samples \cite{han2022multimodal}. MoGE \cite{Cao_2023_ICCV} designed local gating and global gating modules for dynamic fusion. DynMM \cite{xue2023dynamic} considered dynamic scene for text-vision-audio modalities and proposed a new gating network for efficient dynamic fusion. \cite{zhang2023provable} identified the correlations between multi-modal fusion and uncertain estimation and comes up with a robust fusion method.


\vspace{0.35em}
\noindent\textbf{Multi-modal Brain Network Fusion.} To fully integrate data from different modalities for better downstream tasks, several methods have been developed for multi-modal brain network fusion. Generally, current fusion methods can be divided into three categories: \textbf{(1) Representation-level fusion} which directly concats or adds multi-modal representations before final prediction, like MMGNN \cite{sebenius2021multimodal}, GBDM \cite{zhang2021deep} and D-mhgcn \cite{wen2024d}. \textbf{(2) Graph-level fusion} which focuses on fusion graph construction. ALNEGAT \cite{chen2022adversarial} designs an Attention-based graph fusion method by learning a fusion adjancency matrix. Cross-GNN \cite{yang2023mapping} proposes a graph-based framework to capture finer cross-modal structure information. \textbf{(3) Transformer-based} fusion. RH-BrainFS  \cite{ye2024rh} uses bottleneck transformer to avoid direct interaction between SC and FC. NeuroPath \cite{wei2024neuropath} introduces coupling module between SC and FC to improve accuracy of FC-based prediction.

\section{Methodology}
\subsection{Notations}
For a given multi-modal brain network dataset, we define it as $\mathcal{D} = \{(\mathcal{G}_{SC}^i, \mathcal{G}_{FC}^i), y_i\}_{i=1}^{n}$, where $n$ represents the sample size. For each sample, $\mathcal{G}_{SC}^i, \mathcal{G}_{FC}^i$ stands for structural connectivity graph and functional connectivity graph respectively, and $y_i$ is the corresponding label (depending on different tasks, gender, disease, etc) and $y_i \in \{0,1\}$. For brain network of each modality: $\mathcal{G}_{m} = (A_m, X_m, \mathcal{V}_{m}, \mathcal{E}_{m}), m \in \{SC, FC\}$, nodes $\mathcal{V}_m$ represents different brain regions in the brain while edges $\mathcal{E}_m$ represents the connectivity between two brain regions. $X_m\in\mathbb{R}^{N\times d}$ denotes node features for each region and $A_m \in \mathbb{R}^{N \times N}$ denotes the adjacency matrix for brain network $\mathcal{G}_m$.

\subsection{Overview of M$^3$D-BFS}
In this section, we will introduce more details about our M$^3$D-BFS. The structure of our framework is shown in Fig. \ref{M3D-BFS} (stage 3), we project uni-modal embeddings and send them into following MoE blocks. Via MoE's gating mechanism, different brain regions can be assigned to following experts dynamically, thus realizing sample-adaptive dynamic fusion. To facilitate better training, we divide our M$^3$D-BFS into 3 stages: \textbf{(1) stage 1: Uni-modal encoder pretraining}, where we train SC and FC encoders respectively. \textbf{(2) stage 2: Single expert pretraining}, where we train our multi-modal fusion model using single expert network for SC/FC/Fusion. \textbf{(3) stage 3: MoE finetuning} for dynamic multi-modal fusion, where we finetune MoE blocks while keeping other modules frozen. 

\subsection{Uni-modal encoder pretraining}
Directly training a multi-modal fusion model can lead to sub-optimal performance, sometimes the performance of the fusion model can be worse than uni-modal models. This phenomenon of insufficient learning during multi-modal fusion is called ``modality laziness'' \cite{du2023uni} which hurts fusion model's generalization ability. 

Thus, encouraged by \cite{du2023uni}, we first train uni-modal encoders for FC/SC respectively, which \textbf{serve as guidance for multi-modal encoder in stage 2}. Formally, for given data $\{(\mathcal{G}_{FC}^i, \mathcal{G}_{SC}^i), y_i\}$ which stands for $i\mbox{-}th$ sample, we denote the uni-modal encoder as:
\begin{small}
\begin{equation}
    \hat{y_i}_m = g_{\theta_m}\circ f_{\varphi_m}(\mathcal{G}_{m}^i) = g_{\theta_m}(f_{\varphi_m}(\mathcal{G}_{m}^i))
    \label{uni-enc}
\end{equation}
\end{small}
Here $m \in \{SC,FC\}$. $f_{\varphi_m}$ denotes the encoder and here we realize it with a two-layer GCN. $g_{\theta_m}$ is for classifier and we use a one-layer MLP (multi-layer perceptron) for implementation. For either SC or FC, we use cross-entropy loss ($\mathcal{L}_{ce}$) to optimize the uni-modal model. 

After uni-modal training, we save the weights for SC/FC encoders ($f_{\varphi_{SC}}, f_{\varphi_{FC}}$), which will be used for soft alignment in stage 2.

\subsection{Single expert pretraining}
We train our multi-modal fusion model after we pretrain uni-modal encoders. However, directly training MoE is challenging in practice, which may lead to instability during training process \cite{shazeer2017outrageously}. 

According to \cite{li2024uni,lin2024moe,chen2024eve}, progressive training strategy is an effective way to train powerful MoE models. To this end, we further adopt a \textbf{``pretrain then finetune'' paradigm} for MoE training \cite{li2024uni}. Formally, we first train single expert then finetune the MoEs. M$^3$D-BFS is independent of the specific architecture of MoE experts and here we utilize MLP as our expert network. For representations from SC/FC encoders (denote as $h_{SC}, h_{FC}$), we project and feed them into following experts.  Corresponding to our design of MoEs, we consider three types of expert MLPs: FC-specific, SC-specific and Fusion experts. Modal-specific experts take the representation of uni-modal as input while Fusion expert takes the concatenation of SC and FC. $\textbf{READOUT}(\cdot)$ will be applied for the output of each expert before concatenation and final prediction (shown in Eq. \ref{equ_stage2}, and $\textbf{FFN}(\cdot)$ is the feedforward network layer).
\begin{small}
\begin{equation}
\begin{aligned}
\begin{cases}
    \label{equ_stage2}
    \hat{y_i} &= \textbf{FFN}(\text{Concat}(z_{SC}^i , \ z_{FC}^i , \ z_{Fusion}^i))
    \\
    z^i_m &= \textbf{READOUT}(h_m^i), m \in \{SC, FC, Fusion\}
\end{cases}
\end{aligned}
\end{equation}
\end{small}

However, simple cross-entropy loss ($\mathcal{L}_{ce}$) is not enough for multi-modal supervised learning and the reasons lie in label's insufficient supervision for different modalities. Therefore, we add the supervision of pretrained uni-modal encoders from stage 1 \cite{du2023uni}. For representations through encoders $(f_{\phi_{SC}}, f_{\phi_{FC}})$ from stage 1, we design a uni-modal guided distillation loss as follows:
\begin{small}
    \begin{equation}
        \mathcal{L}_{distill} = \frac{\tau^2}{2} \bigg ( \mathcal D_{KL}(\frac{h_{\varphi_{FC}}}{\tau},\frac{h_{\phi_{FC}}}{\tau}) + \mathcal{D}_{KL}(\frac{h_{\varphi_{SC}}}{\tau},\frac{h_{\phi_{SC}}}{\tau})\bigg )
    \end{equation}
\end{small}

The output of the uni-modal encoders from stage 2 are denoted as ($h_{\phi_{SC}}, h_{\phi_{FC}}$) and we regard stage 1 representations as teacher logits $(h_{\varphi_{SC}}, h_{\varphi_{FC}})$. Through $\mathcal{L}_{distill}$ as soft alignment, our \model can keep the performance of uni-modal encoders.

Besides, for better SC/FC representations, we adopt practice from LIMoE \cite{mustafa2022multimodal} pretraining which utilized CLIP-based loss \cite{radford2021learning} between batch data. For given SC/FC representations $(h_{\phi_{SC}}, h_{\phi_{FC}}) \in \mathbb{R}^{N \times d}$, we utilize \underline{\textbf{C}}ontrastive \underline{\textbf{M}}ultimodal \underline{\textbf{B}}rain \underline{\textbf{P}}re-training loss (\textbf{CMBP}, denote as $\mathcal{L}_{contrast}$) for better align representations between SC and FC. We use pooling for $(h_{\phi_{SC}}, h_{\phi_{FC}})$, and get $d$-dimension representations for SC and FC modality.
$\mathcal{L}_{contrast}$ minimizes the distance between paired SC-FC data within one batch during training: 
 $\mathcal{L}_{contrast} = \frac{1}{2}(\mathcal{L}_{S2F} + \mathcal{L}_{F2S})$. Here we use $\textbf{InfoNCE}(\cdot)$ for implementation.

During this stage, classification loss $\mathcal{L}_{ce}$, distillation loss $\mathcal{L}_{distill}$ and $\mathcal{L}_{contrast}$ together make the loss function (Eq. \ref{equ_loss_s2}, where $\beta$ is a hyper-parameter ranging $(0,1)$).
\begin{equation}
    \mathcal{L}_{stage2} = \mathcal{L}_{ce} + \beta\cdot(\mathcal{L}_{distill} + \mathcal{L}_{contrast})
    \label{equ_loss_s2}
\end{equation}

After training, we save the weights of single MLP experts (SC, FC, Fusion), which will be used in stage 3.

\begin{table*}[!htbp]
    \centering
    \caption{Comparison results on two different datasets (HCP, ZDXX). We run 10 times and record average acc $\pm$ std (\%). The best results are marked \textbf{bold}, the second best \underline{underline} and the third best \colorbox{lightgray}{gray}.}
    \resizebox{0.9\textwidth}{!}{
    \begin{tabular}{clcccccc}
        \toprule
        \multirow{1}{*}{\textbf{Datasets}} &
        \multirow{1}{*}{\textbf{Methods}} &
        \multirow{1}{*}{\textbf{Modality}} &
        \textbf{ACC} & \textbf{SEN} & \textbf{SPE} & \textbf{F1} & \textbf{AUC}
        \\
        \midrule
        \multirow{12}{*}{\Large{\textbf{HCP}}}
        & GCN & FC & 64.03 $\pm$ \scriptsize{1.21} & 55.01 $\pm$ \scriptsize{2.88} & 71.75 $\pm$ \scriptsize{2.37} & 57.15 $\pm$ \scriptsize{2.49} & 63.38 $\pm$ \scriptsize{1.25}
        \\
        & GCN & SC & 68.23 $\pm$ \scriptsize{1.76} & 66.22 $\pm$ \scriptsize{4.79} & 72.27 $\pm$ \scriptsize{4.53} & 67.91 $\pm$ \scriptsize{2.71} & 69.61 $\pm$ \scriptsize{1.75}
        \\
        & BrainNPT & FC & 67.78 $\pm$ \scriptsize{1.53} & 60.67 $\pm$ \scriptsize{1.34} & 73.98 $\pm$ \scriptsize{2.83} & 59.15 $\pm$ \scriptsize{3.70} & 66.97 $\pm$ \scriptsize{2.44}
        \\
        & BrainNPT & SC & 70.74 $\pm$ \scriptsize{1.49} & 68.95 $\pm$ \scriptsize{4.82} & 70.02 $\pm$ \scriptsize{5.67} & 68.16 $\pm$ \scriptsize{2.13} & 70.11 $\pm$ \scriptsize{1.73}
        \\
        \cmidrule{2-8}
        & SVM & FC, SC &  74.49 $\pm$ \scriptsize{2.97} & 71.18 $\pm$ \scriptsize{4.95} & 77.32 $\pm$ \scriptsize{3.30} & 71.96 $\pm$ \scriptsize{3.49} & 74.25 $\pm$ \scriptsize{3.05} 
        \\
        & Random Forest & FC, SC & 68.24 $\pm$ \scriptsize{2.94} & 54.88 $\pm$ \scriptsize{5.78} & 79.64 $\pm$ \scriptsize{4.16} & 61.31 $\pm$ \scriptsize{4.44} & 67.26 $\pm$ \scriptsize{3.04}
        \\
        & MMGNN & FC, SC & 73.33 $\pm$ \scriptsize{2.82} & 71.17 $\pm$ \scriptsize{4.52} & 75.17 $\pm$ \scriptsize{5.67} & 71.10 $\pm$ \scriptsize{2.88} & 73.17 $\pm$ \scriptsize{2.72}
        \\
        & AL-NEGAT & FC, SC & 75.12 $\pm$ \scriptsize{3.66} & 72.86 $\pm$ \scriptsize{7.74} & \underline{84.46 $\pm$ \scriptsize{5.05}} & 76.13 $\pm$ \scriptsize{4.70} & \colorbox{lightgray}{78.66 $\pm$ \scriptsize{3.81} }
        \\
        & Cross-GNN & FC, SC & \underline{78.93 $\pm$ \scriptsize{1.05}} & 73.53 $\pm$ \scriptsize{1.84} & \textbf{85.04 $\pm$ \scriptsize{2.02}} & \textbf{76.81 $\pm$ \scriptsize{0.86}} & \underline{79.82 $\pm$ \scriptsize{0.72}}
        \\
        & RH-BrainFS & FC, SC & \colorbox{lightgray}{78.63 $\pm$ \scriptsize{4.36}} & \underline{75.59 $\pm$ \scriptsize{6.75}} & 81.25 $\pm$ \scriptsize{6.04} & \colorbox{lightgray}{76.49 $\pm$ \scriptsize{6.91}} & 78.42 $\pm$ \scriptsize{4.38}
        \\
        & NeuroPath & FC, SC & 77.77 $\pm$ \scriptsize{1.87}  & \colorbox{lightgray}{75.24 $\pm$ \scriptsize{2.61}}  & 80.38 $\pm$ \scriptsize{1.50} & 75.78 $\pm$ \scriptsize{1.84} & 77.81 $\pm$ \scriptsize{1.67}
        \\
        \cmidrule{2-8}
        & \textbf{\model (Ours)} & FC, SC & \textbf{81.21 $\pm$ \scriptsize{0.85}} & \textbf{76.14 $\pm$ \scriptsize{3.32}}  & \colorbox{lightgray}{82.50 $\pm$ \scriptsize{3.07}}  & \underline{76.75 $\pm$ \scriptsize{1.25}} & \textbf{80.80 $\pm$ \scriptsize{1.07}}
        \\

        \midrule\midrule
        \multirow{12}{*}{\Large{\textbf{zhongdaxinxiang}}}
        & GCN & FC & 68.63 $\pm$ \scriptsize{1.28} & 64.61 $\pm$ \scriptsize{4.67} & 72.68 $\pm$ \scriptsize{4.52} & 64.68 $\pm$ \scriptsize{3.13} & 68.64 $\pm$ \scriptsize{1.18}
        \\
        & GCN & SC & 72.15 $\pm$ \scriptsize{2.49} & 72.79 $\pm$ \scriptsize{5.14} & 73.24 $\pm$ \scriptsize{4.80} & 71.90 $\pm$ \scriptsize{4.25} & 72.02 $\pm$ \scriptsize{2.52}
        \\
        & BrainNPT & FC & 70.65 $\pm$ \scriptsize{1.85} & 73.00 $\pm$ \scriptsize{4.59} & 69.01 $\pm$ \scriptsize{4.30} & 69.27 $\pm$ \scriptsize{2.64} & 71.00 $\pm$ \scriptsize{2.03}
        \\
        & BrainNPT & SC & 72.83 $\pm$ \scriptsize{1.73} & 74.57 $\pm$ \scriptsize{3.80} & 69.89 $\pm$ \scriptsize{3.87} & 73.63 $\pm$ \scriptsize{4.07} & 72.78 $\pm$ \scriptsize{2.65}
        \\
        \cmidrule{2-8}
        & SVM & FC, SC &  66.06 $\pm$ \scriptsize{1.56} & 58.01 $\pm$ \scriptsize{2.49} & 74.04 $\pm$ \scriptsize{1.77} & 62.03 $\pm$ \scriptsize{2.27} & 66.03 $\pm$ \scriptsize{1.56} 
        \\
        & Random Forest & FC, SC & 62.43 $\pm$ \scriptsize{2.19} & 56.17 $\pm$ \scriptsize{2.87} & 68.60 $\pm$ \scriptsize{3.48} & 59.04 $\pm$ \scriptsize{2.73} & 62.38 $\pm$ \scriptsize{2.20}
        \\
        & MMGNN & FC, SC & 59.72 $\pm$ \scriptsize{3.18} & 65.10 $\pm$ \scriptsize{4.83} & 54.42 $\pm$ \scriptsize{4.42} & 59.96 $\pm$ \scriptsize{4.02} & 59.76 $\pm$ \scriptsize{3.15}
        \\
        & AL-NEGAT & FC, SC & 71.86 $\pm$ \scriptsize{2.49} & 75.26 $\pm$ \scriptsize{3.62} & 68.12 $\pm$ \scriptsize{6.19} & 72.16 $\pm$ \scriptsize{2.24} & 71.69 $\pm$ \scriptsize{2.56}
        \\
        & Cross-GNN & FC, SC & 74.01 $\pm$ \scriptsize{2.33} & \textbf{77.39 $\pm$ \scriptsize{5.71}} & 70.46 $\pm$ \scriptsize{5.65} & 74.38 $\pm$ \scriptsize{3.15} & 73.92 $\pm$ \scriptsize{2.33}
        \\
        & RH-BrainFS & FC, SC & \underline{78.48 $\pm$ \scriptsize{1.43}} & 76.20 $\pm$ \scriptsize{4.06} & \underline{80.72 $\pm$ \scriptsize{3.60}} & \colorbox{lightgray}{77.35 $\pm$ \scriptsize{1.97}} & \colorbox{lightgray}{78.46 $\pm$ \scriptsize{1.43}}
        \\
        & NeuroPath & FC, SC & \colorbox{lightgray}{78.01 $\pm$ \scriptsize{1.17}} & \colorbox{lightgray}{77.03 $\pm$ \scriptsize{2.23}} & \colorbox{lightgray}{79.17 $\pm$ \scriptsize{3.47}} & \underline{77.54 $\pm$ \scriptsize{2.28}} & \underline{78.50 $\pm$ \scriptsize{1.24}}
        \\
        \cmidrule{2-8}
        & \textbf{\model (Ours)} & FC, SC & \textbf{80.85 $\pm$ \scriptsize{2.44}} & \underline{77.22 $\pm$ \scriptsize{3.04}} & \textbf{83.91 $\pm$ \scriptsize{4.95}} & \textbf{77.82 $\pm$ \scriptsize{3.62}} & \textbf{80.74 $\pm$ \scriptsize{2.57}}
        \\

        \bottomrule
    \end{tabular}    
    } 
    \label{table_comp_HCP}
\end{table*}

\subsection{MoE finetuning}
\noindent\textbf{MoE finetuning for dynamic multi-modal fusion. }Based on the pretrained model in stage 2, We finally finetune our \model in stage 3 to make our training process more stable. In this stage we replace the MLP experts with MoEs, whose experts have the same structure as MLP experts. As is shown in Fig. \ref{M3D-BFS}, We only train MoEs during this stage while keeping other modules frozen. We consider three types of MoEs: Uni-modal MoEs (SC-specific, FC-specific) which capture SC-only or FC-only features and Multi-modal MoE which dynamically fuses SC and FC features. Moreover, to avoid insufficient learning for some experts in MoE training, we apply the MLP weights saved in stage 2 to initialize all the expert networks in corresponding MoEs. 

Each MoE consists of two parts: expert networks and gating network. Here we follow the classical practice from \cite{shazeer2017outrageously} and use MLP to realize both modules. For MoE $m$ which denotes its type ($m \in \{\ SC, FC, Fusion\}$), the output can be written as follows:
\begin{small}
\begin{equation}
    y_m = \sum_{i=1}^{N_m} G_m(h_m)_i \cdot E_m(h_m)_i
\end{equation}
\end{small}

Where the input $h_m$ takes the dimension of $\mathbb{R}^{N\times d}$ for $m \in \{SC, FC\}$ ($N_m=N$) and $\mathbb{R}^{2N\times d}$ for $m\in \{Fusion\}$ ($N_m=2N$). $G_m$ is gating network for corresponding MoE. Following \cite{shazeer2017outrageously}, we apply \textbf{Softmax} function for gating output. Before taking softmax, we add Gaussian noise and only keep top k values. Here we set k as 1, which means each brain region selects only one expert during training and inference. As the output of $G_m$ depends on input $X$, it can dynamically assign brain regions to different experts according to different data samples.

However, direct training process can get collapsed easily, where all input can be assigned to only one expert, and distributions of gating network can be imbalanced, where favored experts are trained more rapidly and are preferred during inference while other experts will get insufficient training \cite{chi2022representation,shazeer2017outrageously}. Thus MoE will degenerate into a single expert. To alleviate this issue of imbalanced training between experts, we apply the Importance loss and Load loss for each MoE ($\mathcal{L}_{moe}$). For given MoE, the loss is the combination of $\mathcal{L}_{Importance}$ and $\mathcal{L}_{load}$,
where the importance loss and load loss are defined as:
\begin{small}
\begin{equation}
\begin{cases}
\begin{aligned}
    \mathcal{L}_{Importance}(X) &= CV(Importance(X))^2 = CV(\sum_{x\in X}G(x))^2
    \\
    \mathcal{L}_{Load}(X) &= CV(Load(X))^2 = CV(\sum_{x\in X}P(x))^2
    \label{equ_moe}
\end{aligned}
\end{cases}
\end{equation}
\end{small}
Here $CV(\cdot)$ means the coefficient of variation of the values. The initialization of experts' weights, as well as the MoE loss function together alleviate imbalanced training issue of MoE which makes the training more stable and more efficient.


\vspace{0.35em}
\noindent\textbf{Representation-enhanced multi-modal disentanglement loss.} In multi-modal learning, insufficient label information for supervision may influence the performance for modal-specific and fusion embeddings. One solution is through disentangled representation learning which reduces redundancy between modal-specific and modal-fusion information while keeping their correlations with label. Here, derived from \cite{li2022modeling}'s method, we design our multi-mdoal disentanglement loss to learn better uni-modal and multi-modal representations ($\mathcal{L}_{disen}$). We apply pooling function after the output of MoE and for each data, the final embeddings are $\{z_{SC}, z_{FC}, z_{Fusion}\} \in \mathbb{R}^d$. Our $\mathcal{L}_{disen}$ is based on contrastive loss and we hope that the final embeddings can be as orthogonal to each other as possible for better representation. We use the concatenation of three representations and get the anchor sample for $i\mbox{-}th$ data $z^i_{anchor}$ by concating $ z^i_{SC}, z^i_{FC}$ and $z^i_{Fusion}$. 
$z^i_{anchor} \in \mathbb{R}^d$ is used for prediction and for each $z_m^i$, we regard other types as negative samples and $z^i_{anchor}$ as positive sample. 
$\mathcal{L}_{disen}$ can be written as Eq. \ref{equ_loss_disen}:
\begin{small}
\begin{equation}
\begin{aligned}
&\mathcal{L}_{disen} =  -\frac{1}{n\cdot M}\sum_{i=1}^{n}\sum_{m=1}^{M}
\\
&  \quad 
\log
\frac{
\exp 
    \big(
        \frac{z_m^i \cdot z_{anchor}^{i}}{\tau_d} 
    \big)
}
{
\exp\big(
        \frac{z_m^i \cdot z_{anchor}^{i}}{\tau_d} 
    \big) + 
    \sum_{\substack{q=1, \\ q \not = m}}^{M} \exp \big(
                                    \frac{z_m^i \cdot z_q^i}{\tau_d}
                                    \big)
}
\label{equ_loss_disen}
\end{aligned}
\end{equation}
\end{small}

Here $M$ defers to different types ($\{SC, FC, Fusion\}$). According to Eq. \ref{equ_loss_disen}, through $\mathcal{L}_{disen}$, we can disentangle different representations while increasing their relevance to downstream tasks, thus enhancing their representations.
Finally, the loss function in stage 3 is in Eq. \ref{equ_loss_stage3} with $\alpha \in (0,1)$:

\begin{small}
\begin{equation}
    \mathcal{L}_{stage3} = \mathcal{L}_{ce} + \alpha \cdot \mathcal{L}_{moe} + (1-\alpha)\cdot \mathcal{L}_{disen}
    \label{equ_loss_stage3}
\end{equation}
\end{small}



\section{Experiments}


\subsection{Experimental settings}
\textbf{Datasets.} We evaluate our \model on two different datasets: \textbf{(1) Human Connectsome Project (HCP)} \cite{HCP}, a public dataset on gender classification which consists of 560 female and 479 male samples. \textbf{(2) Major Depressive Disorder dataset from hospitals (zhongdaxinxiang, also short as ZDXX)}, including the Affiliated Zhongda Hospital of Southeast University and  the Second Affiliated Hospital of Xinxiang Medical University, which consists of 94 controls and 93 MDD patients.

We also introduce preprocess of our dataset. \textbf{(1) SC}: The brain's diffusion toolbox of FMRIB is utilized dMRI preprocessing \cite{FMRIB}. Then the brain regions are obtained ($\mathcal{V}_{SC}$) through Anatomical Automatic labeling (AAL) template. After that, we use DSI Studio \cite{DSI} for fiber tracking before we finally get the features of SC graph ($X_{SC}$) by counting the number of structural connective fibres between different regions of AAL. The construction of $A_{SC}$ is based on $X_{SC}$ by adding a certain threshold. \textbf{(2) FC}: The Data Preprocessing Assistant for Resting-State Function (DPARSF) MRI tookit \cite{DPARSF} is utilized for fMRI preprocessing. Then the average time series are computed for each brain region with AAL template. Pearson correlation is then calculated as function matrix, which denotes the feature matrix for FC ($X_{FC}$). Its $A_{FC}$ is obtained by thresholding a certain proportional quantization.

\begin{figure}
    \centering
    \includegraphics[width=0.96\linewidth]{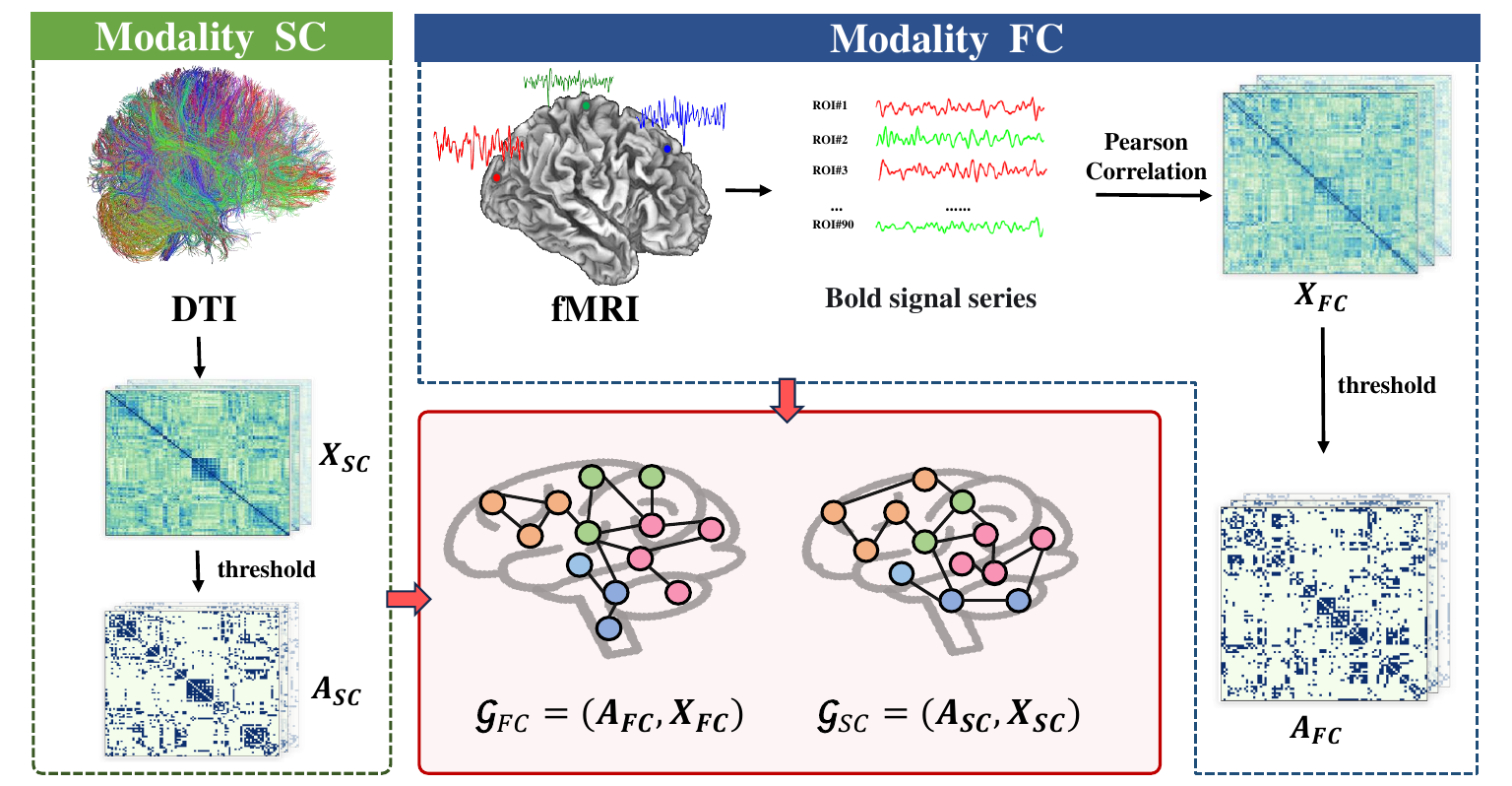}
    \caption{Multi-modal brain networks construction.}
    \label{pdf_preprocess}
\end{figure}




\vspace{0.35em}
\noindent\textbf{Baselines.} 
For uni-modal methods, as we choose \textbf{GCN} as our uni-modal encoders, we record its results in uni-modal scenes. Besides, we select \textbf{BrainNPT} \cite{hu2024brainnpt}, a transformer-based method for brain network analysis. For multi-modal methods, we select (1) 2 methods which concatenate SC/FC features as input: \textbf{SVM}, \textbf{Random Forest}; (2) 1 method concatenates uni-modal embeddings: \textbf{MMGNN} \cite{sebenius2021multimodal}; (3) 2 methods which construct a graph between different modalities for fusion: \textbf{AL-NEGAT} \cite{chen2022adversarial}, \textbf{Cross-GNN} \cite{yang2023mapping} and (4) 2 transformer-based fusion methods: \textbf{RH-BrainFS} \cite{ye2024rh} and \textbf{NeuroPath} \cite{wei2024neuropath}. Code implementations of all baseline methods are taken from their respective original papers. 

\vspace{0.35em}
\noindent\textbf{Experimental settings.} We choose 3-layer GCN with its hidden dimension 128 for uni-modal encoders. For fusion parts, we utilize a two-layer MoE. Top-k value of each MoE block is set as 1. We adopt Adam as optimizer and training epochs is 500 in total with 300 epochs as early stopping. For hyper-parameters settings, $\alpha \in \{0.2, 0.4, 0.6, 0.8\}, \beta \in \{0.3, 0.5, 0.7, 0.9\}$. The number of experts in MoE $k \in [2, 6]$. The learning rate is 0.005 with weight dacay as 0.0001. Our model is implemented in PYG and trained on RTX Titans with 24GB memory. 

We evaluate model's performance on five metrics: accuracy (\textbf{ACC}), sensitivity (\textbf{SEN}), specificity (\textbf{SPE}), f1 score (\textbf{F1}) and ROC-AUC (\textbf{AUC}), where higher value means better performance. We use 10-fold cross validation on 10 random runs and record mean value and standard deviation.

\begin{table}[!htbp]
    \centering
    \caption{Ablation studies on sub-stages and sub-modules.}
    \resizebox{\linewidth}{!}{
    \begin{tabular}{c|cccc|cc}
        \toprule
        \multirow{2}{*}{\textbf{Ablation types}} &
        \multicolumn{1}{c|}{\textbf{Stage 1}} & 
        \multicolumn{2}{c|}{\textbf{Stage 2}} & 
        \multicolumn{1}{c|}{\textbf{Stage 3}} &
        \multicolumn{2}{c}{\textbf{Datasets}}
        \\
         & \multicolumn{1}{c|}{\small\textit{KD}}  & \small\textit{pretrain} & \multicolumn{1}{c|}{\small$\mathcal{L}_{contrast}$} & \small$\mathcal{L}_{disen}$ & HCP & zhongdaxinxiang
        \\
        
        \midrule
        $S_1.A_1$ & \XSolidBrush & \XSolidBrush & \XSolidBrush & \XSolidBrush & 76.84 $\pm$ \scriptsize{1.92} & 77.28 $\pm$ \scriptsize{4.71}\\
        $S_2.A_1$ & \Checkmark & \XSolidBrush & \XSolidBrush & \XSolidBrush & 78.16 $\pm$ \scriptsize{1.57} & 79.67 $\pm$ \scriptsize{4.42}\\
        $S_2.A_2$ & \Checkmark & \Checkmark & \XSolidBrush & \XSolidBrush & 78.57 $\pm$ \scriptsize{1.06} & 77.00 $\pm$ \scriptsize{2.52}\\
        $S_3.A_1$ & \Checkmark & \Checkmark & \Checkmark & \XSolidBrush & 77.80 $\pm$ \scriptsize{1.22} & 78.68 $\pm$ \scriptsize{4.06}\\
        \midrule
        \textbf{M$^3$D-BFS} & \Checkmark & \Checkmark & \Checkmark & \Checkmark & \textbf{81.21 $\pm$ \scriptsize{0.85}} & \textbf{80.85 $\pm$ \scriptsize{2.44}}\\
        \bottomrule
    \end{tabular}    
    } 
    \label{tab_ablation}
\end{table}


\begin{figure}[H]
    \centering
    \includegraphics[width=0.47 \linewidth]{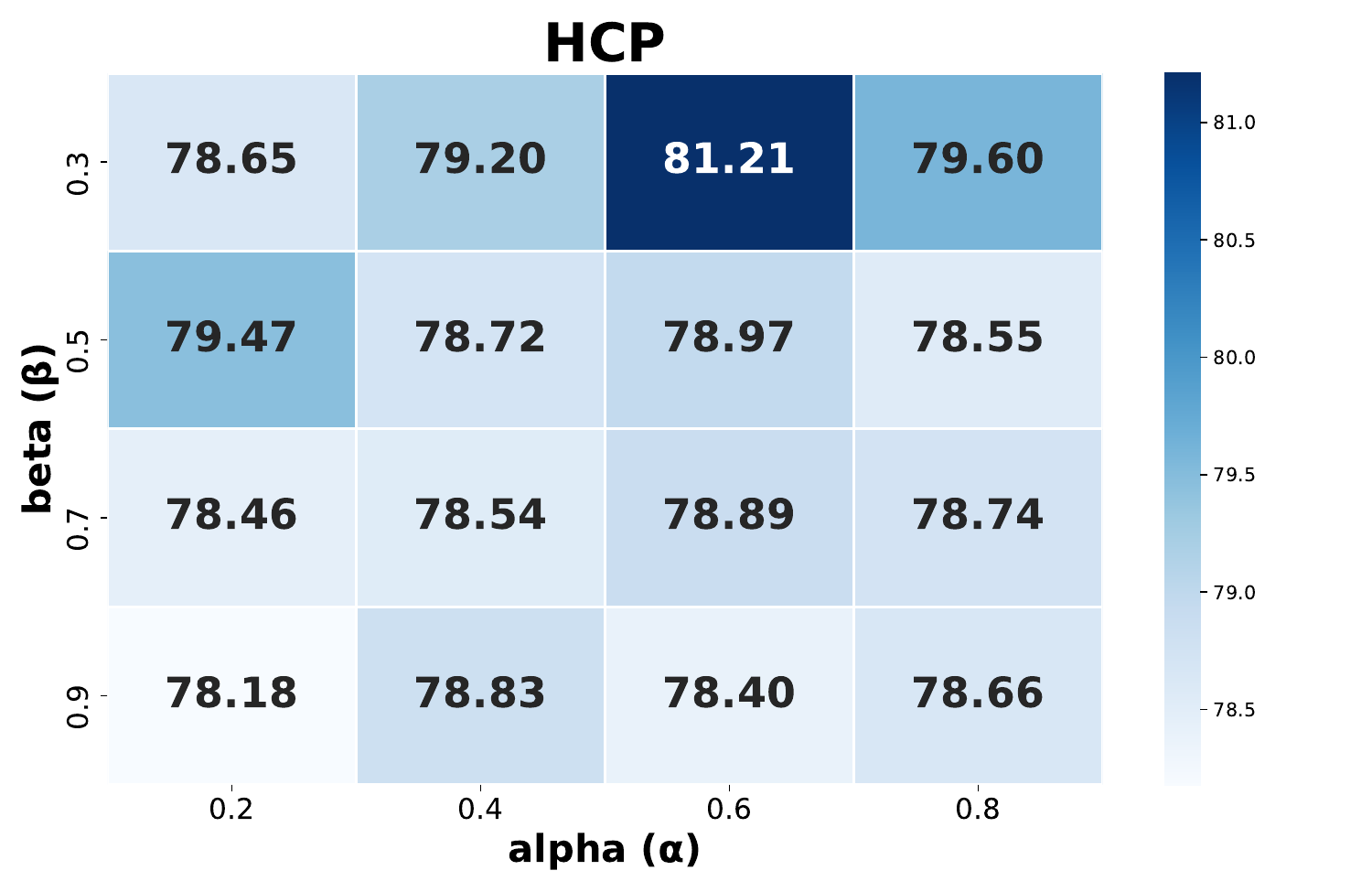}
    \includegraphics[width=0.47\linewidth]{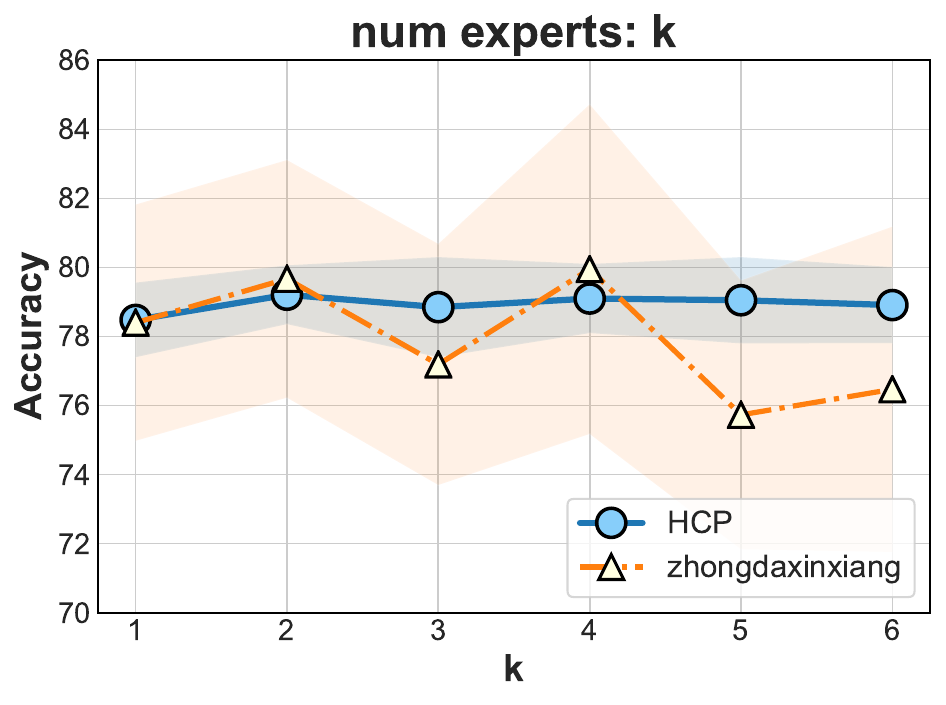}
    
    \caption{Sensitivity analysis experiments.}
    \label{fig_hyper}
\end{figure}

\begin{figure}[H]
    \centering
    \includegraphics[width=0.43\linewidth]{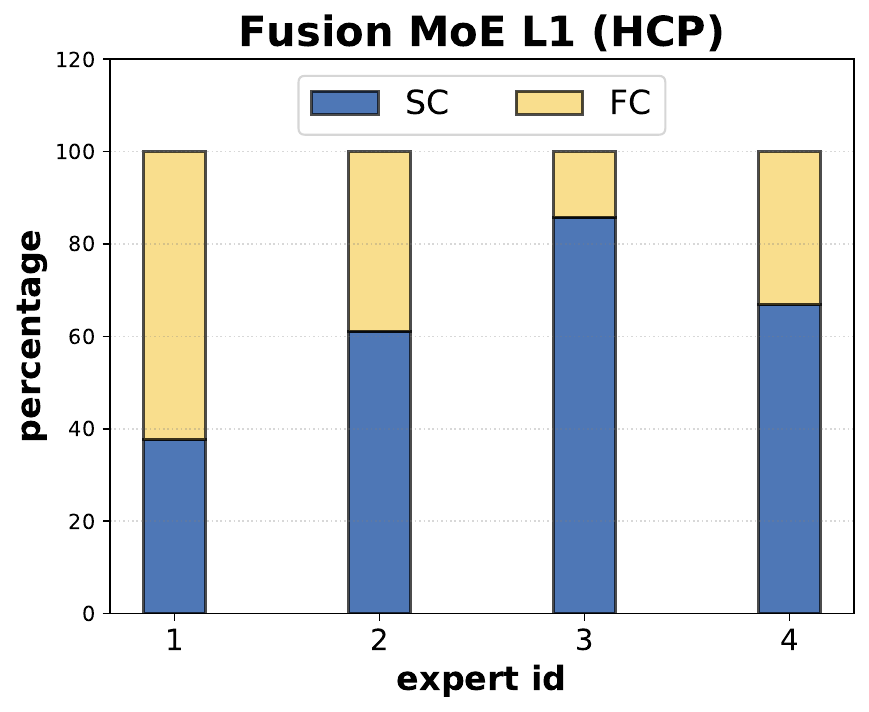}
    \includegraphics[width=0.43\linewidth]{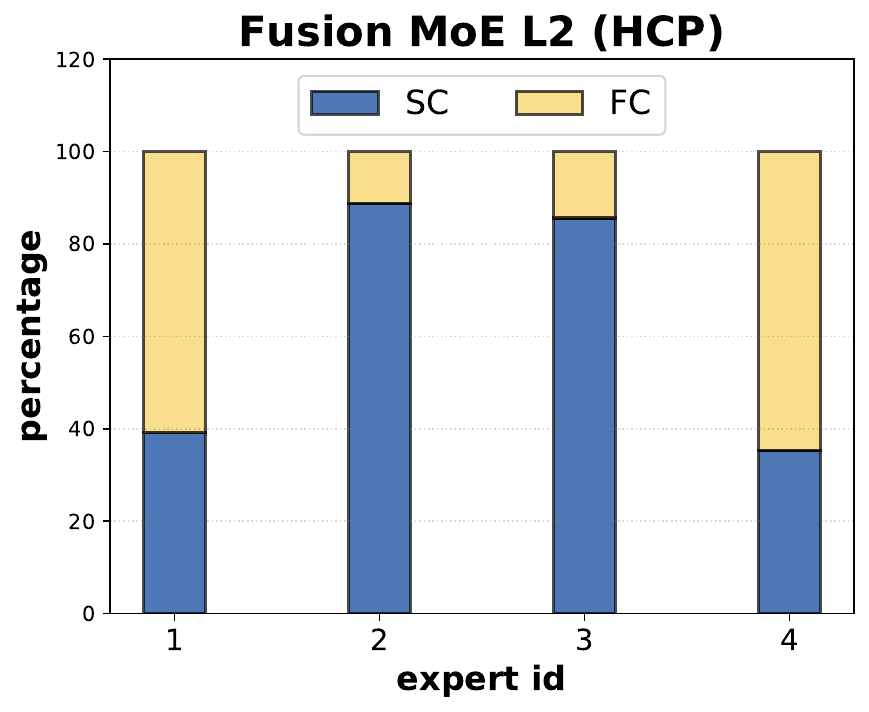}
    \\
    \includegraphics[width=0.43\linewidth]{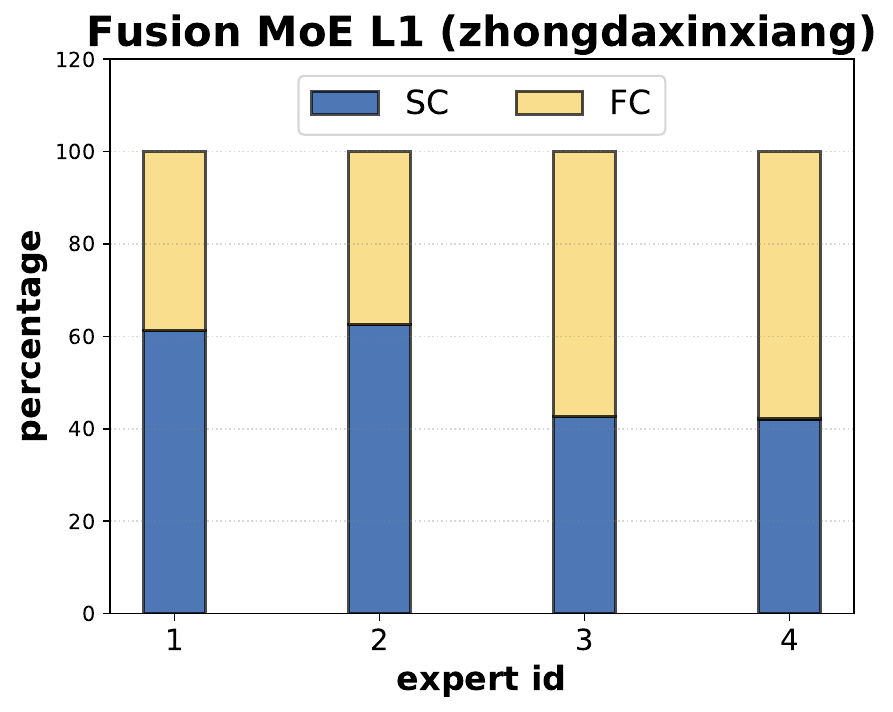}
    \includegraphics[width=0.43\linewidth]{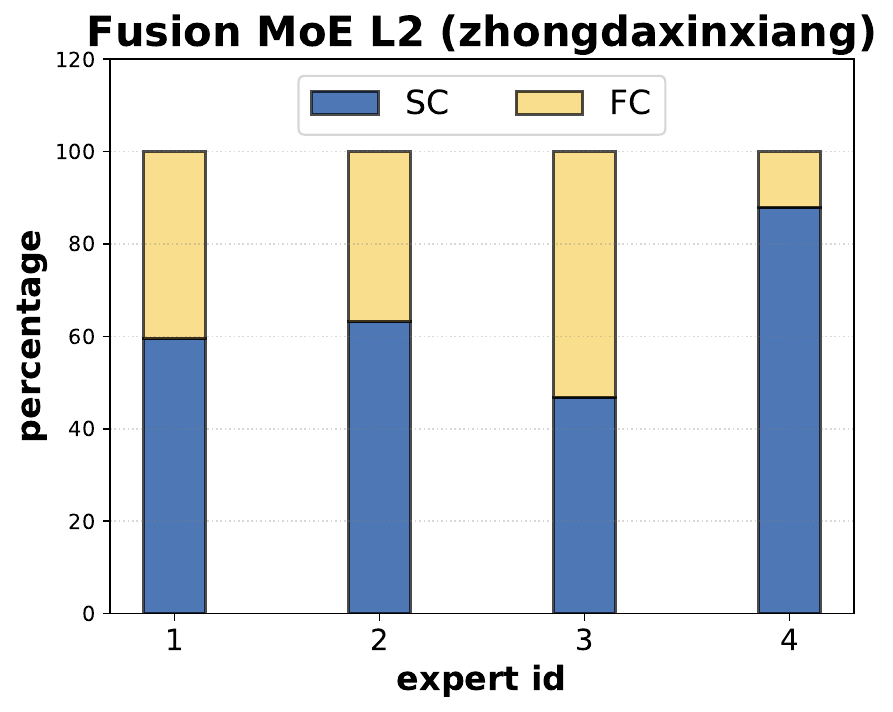}
    
    \caption{Visualization for uni-modal percentage (SC:FC) in each fusion MoE block.}
    \label{fig_moe_fusion}
\end{figure}

\begin{figure*}[!htbp]
    \centering
    \begin{subfigure}[b]{0.22\linewidth}
        \includegraphics[width=\linewidth]{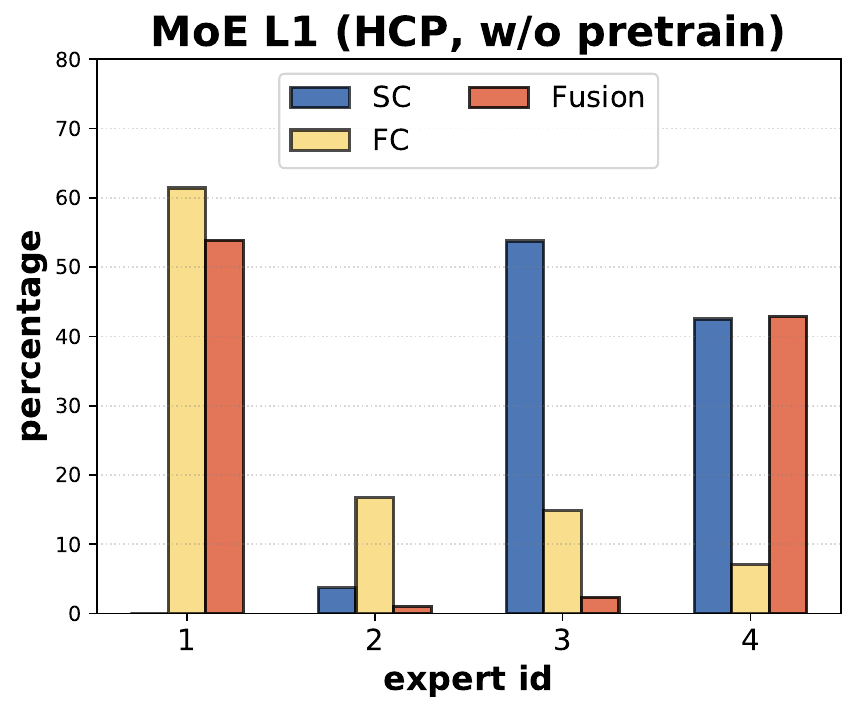} \caption{} \label{a}
    \end{subfigure}
    \begin{subfigure}[b]{0.22\linewidth}
        \includegraphics[width=\linewidth]{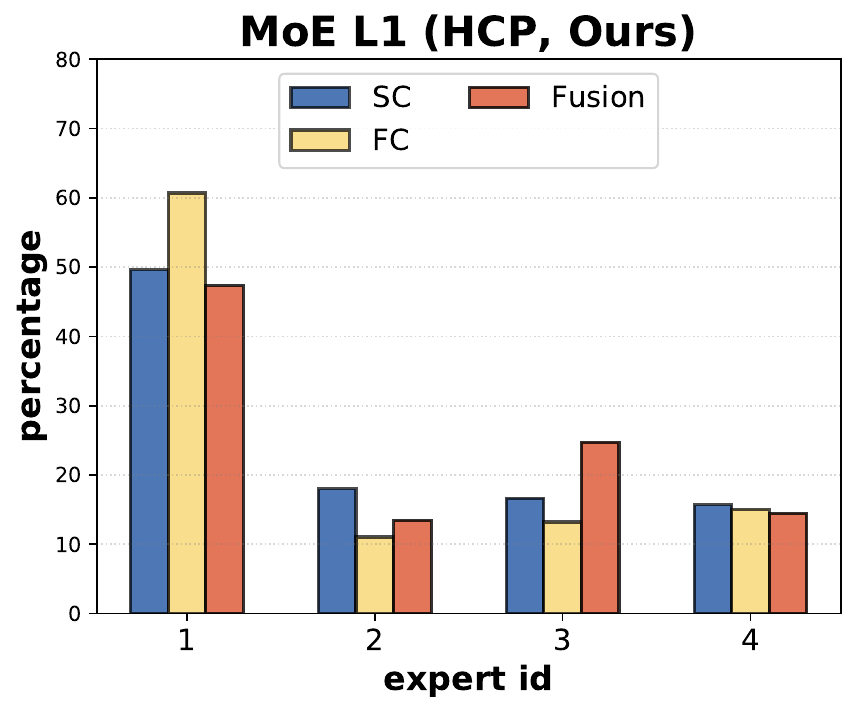} \caption{} \label{b}
    \end{subfigure}
    \begin{subfigure}[b]{0.22\linewidth}
        \includegraphics[width=\linewidth]{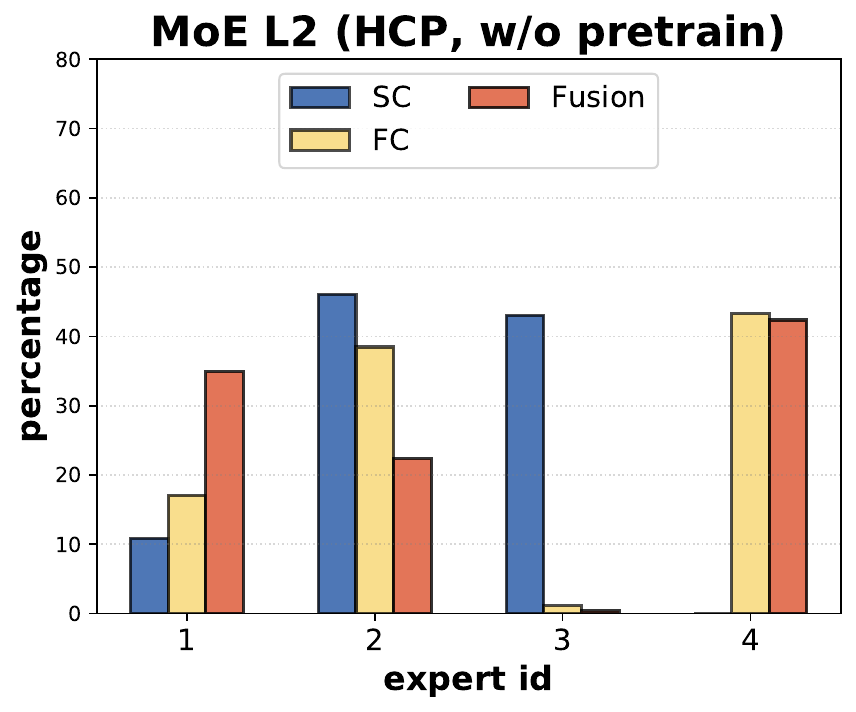} \caption{} \label{c}
    \end{subfigure}
    \begin{subfigure}[b]{0.22\linewidth}
        \includegraphics[width=\linewidth]{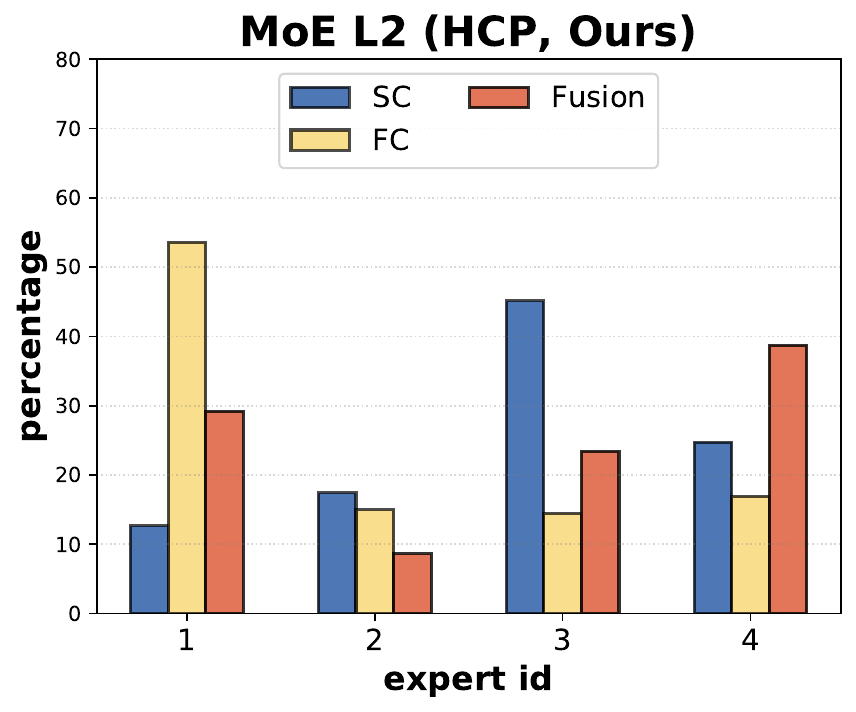} \caption{} \label{d}
    \end{subfigure}

    \begin{subfigure}[b]{0.22\linewidth}
        \includegraphics[width=\linewidth]{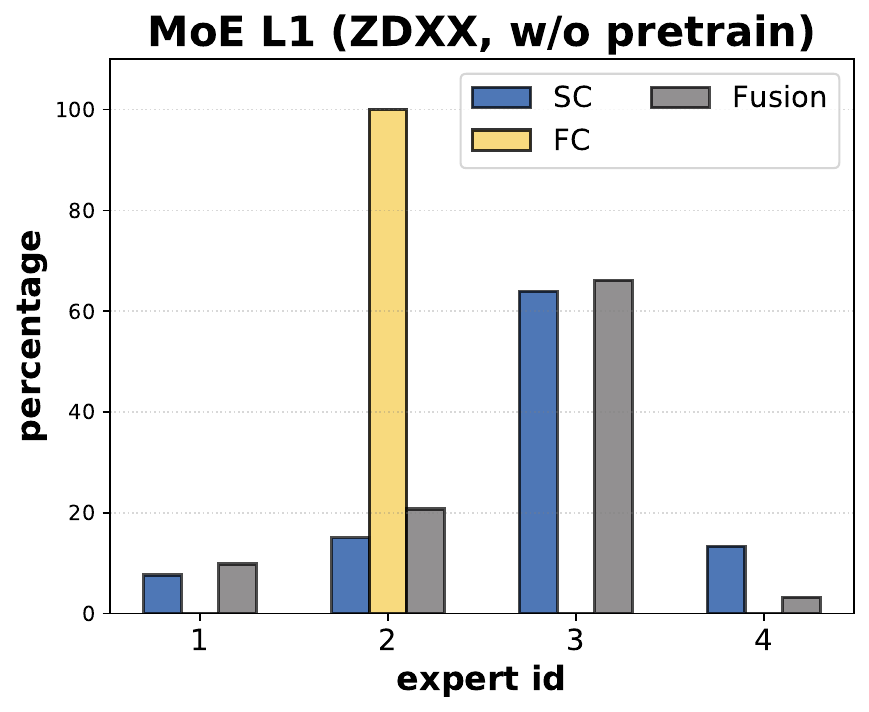} \caption{} \label{e}
    \end{subfigure}
    \begin{subfigure}[b]{0.22\linewidth}
        \includegraphics[width=\linewidth]{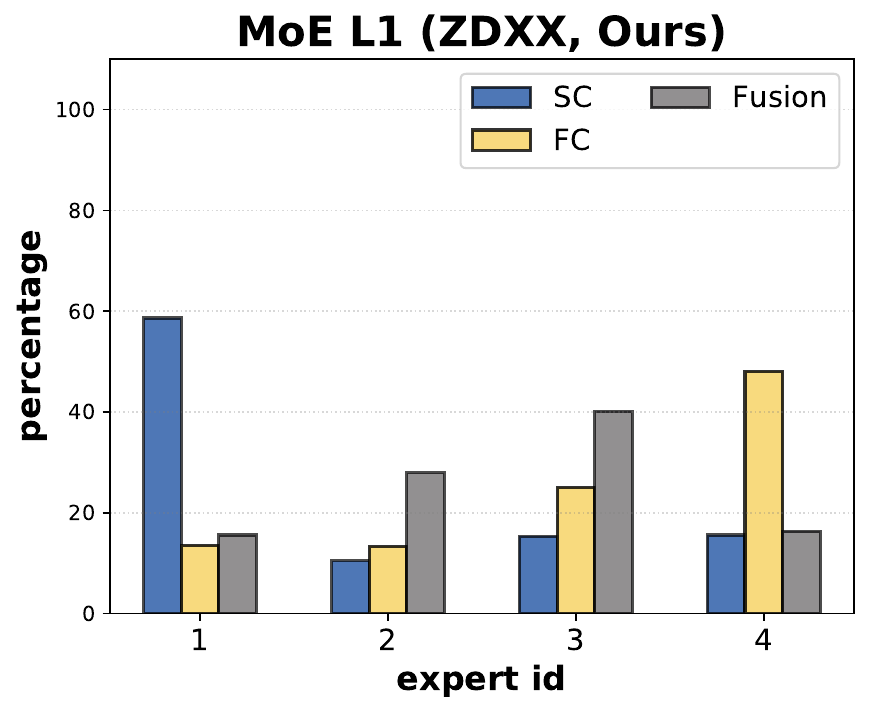} \caption{} \label{f}
    \end{subfigure}
    \begin{subfigure}[b]{0.22\linewidth}
        \includegraphics[width=\linewidth]{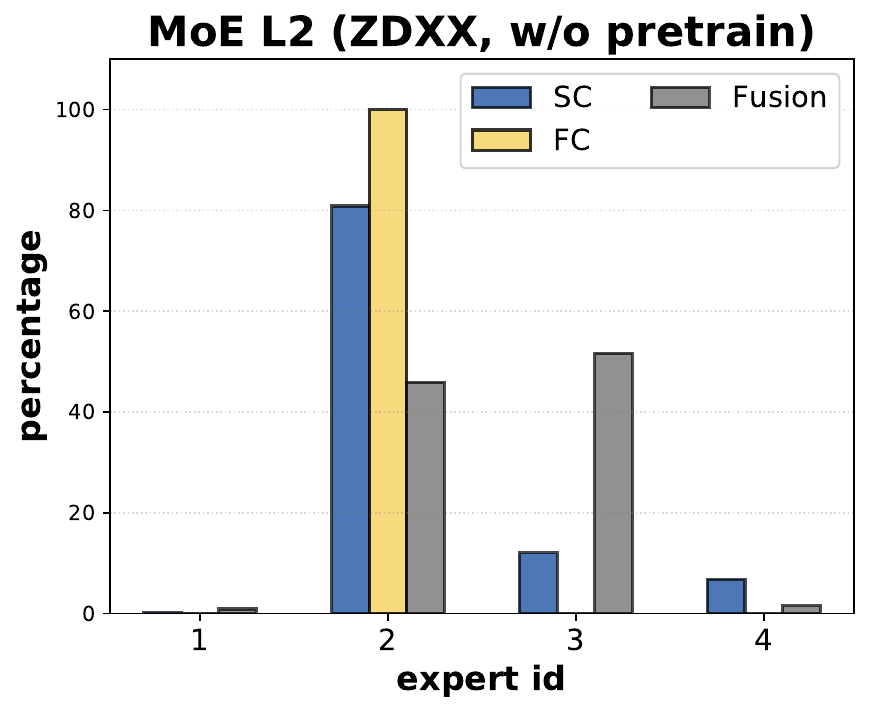} \caption{} \label{g}
    \end{subfigure}
    \begin{subfigure}[b]{0.22\linewidth}
        \includegraphics[width=\linewidth]{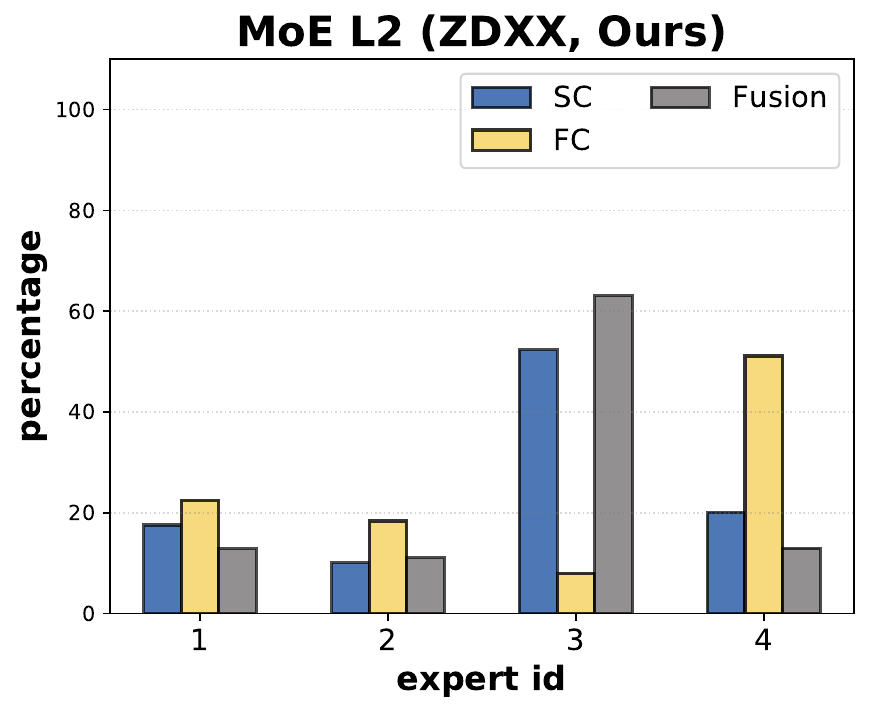} \caption{} \label{h}
    \end{subfigure}

    \caption{Visualization for percentage of experts in different MoEs. Here each MoE consists of 4 experts.}
    \label{fig_moe_all}
\end{figure*}

\subsection{Comparison experiments}
The results of different methods for two datasets are shown in Tab. \ref{table_comp_HCP}. Compared to uni-modal GCNs, \model greatly outperforms uni-modal encoders.
Moreover, compared to other multi-modal brain fusion methods, \model also performs the best (2.28\% \textbf{ACC} improvement on HCP and 2.73\% \textbf{ACC} improvement on ZDXX). 

\subsection{Ablation studies and sensitivity analyses}
We conduct ablation studies to illustrate effectiveness of different sub-modules in M$^3$D-BFS. 
We validate its effectiveness via stage-by-stage ablation. As is shown in Tab. \ref{tab_ablation}, $S_i.A_j$ means ablation for stage $i$ and $j\mbox{-}th$ sub-module in stage $i$. The results in Tab. \ref{tab_ablation} show that each sub-module works in our M$^3$D-BFS. The distillation from uni-modal encoders ensure the models' representations for uni-modal during fusion ($S_1.A_1$). The ``pretrain-finetune'' method not only improves accuracy, but also lowers deviation ($S_2.A_1, S_2.A_2$). Finally $\mathcal{L}_{disen}$ leads to better performance ($S_3.A_1$). 

We also conduct extensive experiments on different hyper-parameters. 
For loss function coefficients, here we consider $\beta$ in stage 2 and $\alpha$ in stage 3 and plot the heatmap for them, which is depicted in Fig. \ref{fig_hyper}. For HCP, the accuracy reaches its peak when the values are $(\alpha, \beta) = (0.6, 0.3)$.
Finally we conduct experiments on different expert numbers in the MoE block. The value of number k is from 1 to 6, and as is shown in Fig. \ref{fig_hyper}, for HCP, the change of accuracy is subtler than ZDXX dataset.


\subsection{Visualization for experts}
Firstly we plot percentage of different modalities in each fusion MoE block. As is shown in Fig. \ref{fig_moe_fusion}, every fusion expert avoid collapse (SC/FC 100\% or 0\%) phenomenon, showing that \model \textbf{gets effective training in dynamic fusion}. Then we plot the distribution of all experts in one MoE block to check if expert collapse occurs in our model. We also plot distributions for models without pretraining (left side) in stage 2 as comparison, which is shown in Fig. \ref{fig_moe_all}. For MoE in both layers, model without pretraining may face collapse among experts. While for M$^3$D-BFS (right side), the issue is effectively mitigated, \textbf{with more even distributions for different experts}.

\section{Conclusion}
 In this work, we propose a multi-stage MoE-based multi-modal brain network fusion method (M$^3$D-BFS) for sample-adaptive dynamic fusion in neuroscience. To the best of our knowledge, this is the first paper discussing dynamic fusion in multi-modal brain fusion. Extensive experiments on different real-world datasets validate the effectiveness of our method. In our future work, we aim to consider more complex cases where different data may have inherent bias in its distribution to achieve better results in different downstream tasks under multi-modal brain network scenarios.


\section*{Acknowledgement}
This work is supported by National Natural Science Foundation of China (Grant No.62471133). 
This work is also supported by the Big Data Computing Center of Southeast University.

{
    \small
    \bibliographystyle{ieeenat_fullname}
    \bibliography{references}

@String(ICCV= {Int. Conf. Comput. Vis.})

@String(AAAI = {AAAI})

@String(ICCV  = {ICCV})

@inproceedings{du2023uni,
  title={On uni-modal feature learning in supervised multi-modal learning},
  author={Du, Chenzhuang and Teng, Jiaye and Li, Tingle and Liu, Yichen and Yuan, Tianyuan and Wang, Yue and Yuan, Yang and Zhao, Hang},
  booktitle={International Conference on Machine Learning},
  pages={8632--8656},
  year={2023},
  organization={PMLR}
}

@article{shazeer2017outrageously,
  title={Outrageously large neural networks: The sparsely-gated mixture-of-experts layer},
  author={Shazeer, Noam and Mirhoseini, Azalia and Maziarz, Krzysztof and Davis, Andy and Le, Quoc and Hinton, Geoffrey and Dean, Jeff},
  journal={arXiv preprint arXiv:1701.06538},
  year={2017}
}

@article{chi2022representation,
  title={On the representation collapse of sparse mixture of experts},
  author={Chi, Zewen and Dong, Li and Huang, Shaohan and Dai, Damai and Ma, Shuming and Patra, Barun and Singhal, Saksham and Bajaj, Payal and Song, Xia and Mao, Xian-Ling and others},
  journal={Advances in Neural Information Processing Systems},
  volume={35},
  pages={34600--34613},
  year={2022}
}

@article{li2024uni,
  title={Uni-MoE: Scaling Unified Multimodal LLMs with Mixture of Experts},
  author={Li, Yunxin and Jiang, Shenyuan and Hu, Baotian and Wang, Longyue and Zhong, Wanqi and Luo, Wenhan and Ma, Lin and Zhang, Min},
  journal={arXiv preprint arXiv:2405.11273},
  year={2024}
}

@article{lin2024moe,
  title={Moe-llava: Mixture of experts for large vision-language models},
  author={Lin, Bin and Tang, Zhenyu and Ye, Yang and Cui, Jiaxi and Zhu, Bin and Jin, Peng and Zhang, Junwu and Ning, Munan and Yuan, Li},
  journal={arXiv preprint arXiv:2401.15947},
  year={2024}
}

@inproceedings{chen2024eve,
  title={Eve: Efficient vision-language pre-training with masked prediction and modality-aware moe},
  author={Chen, Junyi and Guo, Longteng and Sun, Jia and Shao, Shuai and Yuan, Zehuan and Lin, Liang and Zhang, Dongyu},
  booktitle={Proceedings of the AAAI Conference on Artificial Intelligence},
  volume={38},
  number={2},
  pages={1110--1119},
  year={2024}
}

@article{mustafa2022multimodal,
  title={Multimodal contrastive learning with limoe: the language-image mixture of experts},
  author={Mustafa, Basil and Riquelme, Carlos and Puigcerver, Joan and Jenatton, Rodolphe and Houlsby, Neil},
  journal={Advances in Neural Information Processing Systems},
  volume={35},
  pages={9564--9576},
  year={2022}
}

@inproceedings{radford2021learning,
  title={Learning transferable visual models from natural language supervision},
  author={Radford, Alec and Kim, Jong Wook and Hallacy, Chris and Ramesh, Aditya and Goh, Gabriel and Agarwal, Sandhini and Sastry, Girish and Askell, Amanda and Mishkin, Pamela and Clark, Jack and others},
  booktitle={International conference on machine learning},
  pages={8748--8763},
  year={2021},
  organization={PMLR}
}

@article{li2022modeling,
  title={Modeling multiple views via implicitly preserving global consistency and local complementarity},
  author={Li, Jiangmeng and Qiang, Wenwen and Zheng, Changwen and Su, Bing and Razzak, Farid and Wen, Ji-Rong and Xiong, Hui},
  journal={IEEE Transactions on Knowledge and Data Engineering},
  volume={35},
  number={7},
  pages={7220--7238},
  year={2022},
  publisher={IEEE}
}

@article{hu2024brainnpt,
  title={BrainNPT: Pre-training Transformer Networks for Brain Network Classification},
  author={Hu, Jinlong and Huang, Yangmin and Wang, Nan and Dong, Shoubin},
  journal={IEEE Transactions on Neural Systems and Rehabilitation Engineering},
  year={2024},
  publisher={IEEE}
}

@inproceedings{sebenius2021multimodal,
  title={Multimodal graph coarsening for interpretable, MRI-based brain graph neural network},
  author={Sebenius, Isaac and Campbell, Alexander and Morgan, Sarah E and Bullmore, Edward T and Li{\`o}, Pietro},
  booktitle={2021 IEEE 31st International Workshop on Machine Learning for Signal Processing (MLSP)},
  pages={1--6},
  year={2021},
  organization={IEEE}
}

@article{chen2022adversarial,
  title={Adversarial learning based node-edge graph attention networks for autism spectrum disorder identification},
  author={Chen, Yuzhong and Yan, Jiadong and Jiang, Mingxin and Zhang, Tuo and Zhao, Zhongbo and Zhao, Weihua and Zheng, Jian and Yao, Dezhong and Zhang, Rong and Kendrick, Keith M and others},
  journal={IEEE Transactions on Neural Networks and Learning Systems},
  year={2022},
  publisher={IEEE}
}

@article{yang2023mapping,
  title={Mapping Multi-Modal Brain Connectome for Brain Disorder Diagnosis via Cross-Modal Mutual Learning},
  author={Yang, Yanwu and Ye, Chenfei and Guo, Xutao and Wu, Tao and Xiang, Yang and Ma, Ting},
  journal={IEEE transactions on medical imaging},
  volume={43},
  number={1},
  pages={108--121},
  year={2024}
}

@article{ye2024rh,
  title={RH-BrainFS: regional heterogeneous multimodal brain networks fusion strategy},
  author={Ye, Hongting and Zheng, Yalu and Li, Yueying and Zhang, Ke and Kong, Youyong and Yuan, Yonggui},
  journal={Advances in Neural Information Processing Systems},
  volume={36},
  year={2024}
}

@article{wei2024neuropath,
  title={NeuroPath: A Neural Pathway Transformer for Joining the Dots of Human Connectomes},
  author={Wei, Ziquan and Dan, Tingting and Ding, Jiaqi and Wu, Guorong},
  journal={arXiv preprint arXiv:2409.17510},
  year={2024}
}

@article{HCP,
  title={The WU-Minn human connectome project: an overview},
  author={Van Essen, David C and Smith, Stephen M and Barch, Deanna M and Behrens, Timothy EJ and Yacoub, Essa and Ugurbil, Kamil and Wu-Minn HCP Consortium and others},
  journal={Neuroimage},
  volume={80},
  pages={62--79},
  year={2013},
  publisher={Elsevier}
}

@article{FMRIB,
  title={Characterizing the role of the structural connectome in seizure dynamics},
  author={Shah, Preya and Ashourvan, Arian and Mikhail, Fadi and Pines, Adam and Kini, Lohith and Oechsel, Kelly and Das, Sandhitsu R and Stein, Joel M and Shinohara, Russell T and Bassett, Danielle S and others},
  journal={Brain},
  volume={142},
  number={7},
  pages={1955--1972},
  year={2019},
  publisher={Oxford University Press}
}

@article{DPARSF,
  title={REST: a toolkit for resting-state functional magnetic resonance imaging data processing},
  author={Song, Xiao-Wei and Dong, Zhang-Ye and Long, Xiang-Yu and Li, Su-Fang and Zuo, Xi-Nian and Zhu, Chao-Zhe and He, Yong and Yan, Chao-Gan and Zang, Yu-Feng},
  journal={PloS one},
  volume={6},
  number={9},
  pages={e25031},
  year={2011},
  publisher={Public Library of Science San Francisco, USA}
}

@inproceedings{DSI,
  title={Detecting structural brain connectivity differences in dementia through a conductance model},
  author={Frau-Pascual, Aina and Augustinack, Jean and Varadarajan, Divya and Yendiki, Anastasia and Fischl, Bruce and Aganj, Iman},
  booktitle={2019 53rd Asilomar conference on signals, systems, and computers},
  pages={591--595},
  year={2019},
  organization={IEEE}
}

@article{basaia2024brain,
  title={Brain connectivity networks constructed using MRI for predicting patterns of atrophy progression in Parkinson disease},
  author={Basaia, Silvia and Agosta, Federica and Sarasso, Elisabetta and Balestrino, Roberta and Stojkovi{\'c}, Tanja and Stankovi{\'c}, Iva and Tomi{\'c}, Aleksandra and Markovi{\'c}, Vladana and Vignaroli, Francesca and Stefanova, Elka and others},
  journal={Radiology},
  volume={311},
  number={3},
  pages={e232454},
  year={2024},
  publisher={Radiological Society of North America}
}

@article{zhang2021deep,
  title={Deep fusion of brain structure-function in mild cognitive impairment},
  author={Zhang, Lu and Wang, Li and Gao, Jean and Risacher, Shannon L and Yan, Jingwen and Li, Gang and Liu, Tianming and Zhu, Dajiang and Alzheimer’s Disease Neuroimaging Initiative and others},
  journal={Medical image analysis},
  volume={72},
  pages={102082},
  year={2021},
  publisher={Elsevier}
}

@article{wen2024d,
  title={D-MHGCN: An End-to-End Individual Behavioral Prediction Model Using Dual Multi-Hop Graph Convolutional Network},
  author={Wen, Xuyun and Cao, Qumei and Zhao, Yunxi and Wu, Xia and Zhang, Daoqiang},
  journal={IEEE Journal of Biomedical and Health Informatics},
  year={2024},
  publisher={IEEE}
}

@article{duan2017degree,
  title={Degree centrality of the functional network in schizophrenia patients},
  author={Duan, Mingjun and Jiang, Yuchao and Chen, Xi and Luo, Cheng and Yao, Dezhong},
  journal={Sheng Wu Yi Xue Gong Cheng Xue Za Zhi Journal of Biomedical Engineering Shengwu Yixue Gongchengxue Zazhi},
  volume={34},
  number={6},
  pages={837--841},
  year={2017}
}

@article{chou2023functional,
  title={Functional MRI and diffusion tensor imaging in Migraine: a review of Migraine functional and White matter microstructural changes},
  author={Chou, Brendon C and Lerner, Alexander and Barisano, Giuseppe and Phung, Daniel and Xu, Wilson and Pinto, Soniya N and Sheikh-Bahaei, Nasim},
  journal={Journal of Central Nervous System Disease},
  volume={15},
  pages={11795735231205413},
  year={2023},
  publisher={SAGE Publications Sage UK: London, England}
}

@article{bessadok2022graph,
  title={Graph neural networks in network neuroscience},
  author={Bessadok, Alaa and Mahjoub, Mohamed Ali and Rekik, Islem},
  journal={IEEE Transactions on Pattern Analysis and Machine Intelligence},
  volume={45},
  number={5},
  pages={5833--5848},
  year={2022},
  publisher={IEEE}
}

@article{zahid2022brainnet,
  title={BrainNet: optimal deep learning feature fusion for brain tumor classification},
  author={Zahid, Usman and Ashraf, Imran and Khan, Muhammad Attique and Alhaisoni, Majed and Yahya, Khawaja M and Hussein, Hany S and Alshazly, Hammam},
  journal={Computational Intelligence and Neuroscience},
  volume={2022},
  number={1},
  pages={1465173},
  year={2022},
  publisher={Wiley Online Library}
}

@article{winter2024systematic,
  title={A Systematic Evaluation of Machine Learning--Based Biomarkers for Major Depressive Disorder},
  author={Winter, Nils R and Blanke, Julian and Leenings, Ramona and Ernsting, Jan and Fisch, Lukas and Sarink, Kelvin and Barkhau, Carlotta and Emden, Daniel and Thiel, Katharina and Flinkenfl{\"u}gel, Kira and others},
  journal={JAMA psychiatry},
  volume={81},
  number={4},
  pages={386--395},
  year={2024},
  publisher={American Medical Association}
}

@article{yan2019reduced,
  title={Reduced default mode network functional connectivity in patients with recurrent major depressive disorder},
  author={Yan, Chao-Gan and Chen, Xiao and Li, Le and Castellanos, Francisco Xavier and Bai, Tong-Jian and Bo, Qi-Jing and Cao, Jun and Chen, Guan-Mao and Chen, Ning-Xuan and Chen, Wei and others},
  journal={Proceedings of the National Academy of Sciences},
  volume={116},
  number={18},
  pages={9078--9083},
  year={2019},
  publisher={National Acad Sciences}
}

@article{van2020white,
  title={White matter disturbances in major depressive disorder: a coordinated analysis across 20 international cohorts in the ENIGMA MDD working group},
  author={Van Velzen, Laura S and Kelly, Sinead and Isaev, Dmitry and Aleman, Andre and Aftanas, Lyubomir I and Bauer, Jochen and Baune, Bernhard T and Brak, Ivan V and Carballedo, Angela and Connolly, Colm G and others},
  journal={Molecular psychiatry},
  volume={25},
  number={7},
  pages={1511--1525},
  year={2020},
  publisher={Nature Publishing Group UK London}
}

@inproceedings{xue2023dynamic,
  title={Dynamic multimodal fusion},
  author={Xue, Zihui and Marculescu, Radu},
  booktitle={Proceedings of the IEEE/CVF Conference on Computer Vision and Pattern Recognition},
  pages={2575--2584},
  year={2023}
}

@inproceedings{han2022multimodal,
  title={Multimodal dynamics: Dynamical fusion for trustworthy multimodal classification},
  author={Han, Zongbo and Yang, Fan and Huang, Junzhou and Zhang, Changqing and Yao, Jianhua},
  booktitle={Proceedings of the IEEE/CVF conference on computer vision and pattern recognition},
  pages={20707--20717},
  year={2022}
}

@inproceedings{zhang2023provable,
  title={Provable dynamic fusion for low-quality multimodal data},
  author={Zhang, Qingyang and Wu, Haitao and Zhang, Changqing and Hu, Qinghua and Fu, Huazhu and Zhou, Joey Tianyi and Peng, Xi},
  booktitle={International conference on machine learning},
  pages={41753--41769},
  year={2023},
  organization={PMLR}
}

@InProceedings{Cao_2023_ICCV,
    author    = {Cao, Bing and Sun, Yiming and Zhu, Pengfei and Hu, Qinghua},
    title     = {Multi-Modal Gated Mixture of Local-to-Global Experts for Dynamic Image Fusion},
    booktitle = {Proceedings of the IEEE/CVF International Conference on Computer Vision (ICCV)},
    month     = {October},
    year      = {2023},
    pages     = {23555-23564}
}

@article{chu2018function,
  title={Function-specific and enhanced brain structural connectivity mapping via joint modeling of diffusion and functional MRI},
  author={Chu, Shu-Hsien and Parhi, Keshab K and Lenglet, Christophe},
  journal={Scientific reports},
  volume={8},
  number={1},
  pages={4741},
  year={2018},
  publisher={Nature Publishing Group UK London}
}
}


\end{document}